# Extend Commitment Protocols with Temporal Regulations: Why and How

Elisa Marengo, Matteo Baldoni, and Cristina Baroglio*

Dipartimento di Informatica, Università degli Studi di Torino
{emarengo, baldoni, baroglio}@di.unito.it

**Abstract.** The proposal of Elisa Marengo's thesis is to extend commitment protocols to explicitly account for *temporal regulations*. This extension will satisfy two needs: (1) it will allow representing, in a flexible and modular way, temporal regulations with a normative force, posed on the interaction, so as to represent conventions, laws and suchlike; (2) it will allow committing to complex conditions, which describe not only what will be achieved but to some extent also *how*. These two aspects will be deeply investigated in the proposal of a unified framework, which is part of the ongoing work and will be included in the thesis.

## 1 Commitment-based Interaction Protocols

The issues of communication and cooperation are crucial in the area of Multiagent systems (MAS). The common solution is to rely on agent interaction protocols. Among different proposals, *commitment protocols* [24] have been widely adopted. All agents, involved in an interaction ruled by a commitment protocol, share the semantics of a set of actions which affect the social state. This semantics is based on the notion of *social commitment*. The idea is that if an agent takes a commitment towards another agent to bring about a condition, then, it will behave in such a way to fulfill the engagement sooner or later. In this respect, commitment protocols have a *deontic nature*, because a commitment introduces a social expectation on the responsibility of some agent towards some other agent to perform something or to achieve some result. Commitment protocols suit well open MAS because they are *respectful* of the agents' autonomy, since no introspection to the agents' mental states is required [26]; they are *dynamic* because commitments can be created, released, deleted and suchlike, and *flexible* because agents are free to take advantage of opportunities or to follow shortcuts [29].

Commitment protocols have fundamentally changed the process of protocol specification from a procedural approach (i.e., prescribing *how* an interaction is to be executed) to a declarative one (i.e., describing *what* interaction is to take place) [28]. Agents decide which action to perform depending on the commitments they have taken and this is because they want to comply with the protocol and fulfill the engagements they have taken [28]. However, in many practical situations this is not sufficient. *Why*? Because, in many cases it is necessary to express some hints on how the interaction should evolve [7,18]. For example, it is necessary to express that some ways to fulfill the commitments are preferred over others, or that only some of them are legal. This does not mean going back to procedural approaches, but it means reconsidering the *how*. In particular, it could be necessary to express *commitments to temporal regulations* and to represent *legal patterns of interaction*. The former are needed to express the engagement of someone to achieve something and in a specific order. For instance, an insurance company commits to paying an in-network surgeon for a procedure only after a covered patient has undergone the procedure. Patterns of interaction, instead, can capture conventions, laws, preferences, habits, or, in general, rules that hold a given reality. For example, in a democratic assembly, a participant cannot speak if she has not obtained the floor.

---

* M. Baldoni and C. Baroglio are the advisor and co-advisor.

The thesis, therefore, focuses on the *specification of interactions*, which require a degree of expressiveness that commitments alone do not have. We propose an extension of commitment protocols in order to (i) supply a way for expressing patterns of interaction, capturing laws, conventions and whatever constrains the interaction and (ii) extend the regulative nature of commitments with the possibility of explicitly committing to temporal regulations. A further challenge is how to provide a specification of the interaction in which agents can recognize the normative force of the temporal regulations and explicitly accept them. Indeed, since an agent is free to violate or to behave in accordance with a norm, for a regulation to *influence* the agents' behaviour it must be ascribed of a normative force and, then, it must be accepted as a norm by the agent [13]. The final step will investigate this aspect and propose a unified framework in which both patterns of interactions and commitments to temporal regulations can find place.

## 2   Temporal Regulations and Commitment Protocols

We discuss how temporal regulations and commitment protocols can be combined: Section 2.1 describes our proposal for including patterns of interaction in protocol specification; Section 2.2 describes commitments to temporal regulations.

### 2.1   *Why (and How)* expressing legal patterns of interaction

Commitment-based protocols represent a valid solution for interaction protocols specification in open and heterogeneous MAS, mostly because they take into account and respect the agents' and MAS characteristics. For instance, they are respectful of the agents' autonomy, and they do not require a particular implementation or architecture to the agents that are part of the system (heterogeneity). However, to be considered a complete tool for interaction specification, commitment protocols cannot disregard the possibility of specifying *patterns of interaction* as temporal regulations. This requirement is supported by many proposals in the literature (see Section 3) and is due to the need of discriminating those possible executions that are legal from those that are not. These patterns can specify rules of different nature, like habits, conventions, laws, protocol compositions or simply preferences. Basically, they capture a partial ordering between certain actions (or states of affair to be achieved).

Our proposal, described deeply in [4,7], relies on Searle's definition of a social reality [23]. In particular, he identifies a *constitutive* and a *regulative* specification. The former defines a set of actions as foundational of a certain context. In commitment protocols this corresponds to the *social meaning* of the actions, i.e. to actions' semantics, usually given in terms of effects on the social state. The latter, instead, captures how things should be carried on. In current proposals there is not a clear distinction between the constitutive and the regulative part of the specification. Some approaches only care of the regulative nature of commitments (once a commitment is taken, it must be be fulfilled), but completely disregard a specification of *how* things should be done. The thesis proposes an explicit and declarative definition of the regulative specification, given by means of a set of constraints expressed in Linear-time Temporal Logic (LTL). Such constraints [4] define a relative order among different conditions (facts and commitments) that become true in the social state. For example, it allows to express that something can become true only after something else holds, or that if something becomes true, something else must hold sooner or later. The choice of a declarative representation of constraints allows for the specification of what is mandatory and what is forbidden in a protocol, without the need to enumerate the allowed executions (on the contrary to procedural approaches). Indeed, such enumeration is often a huge task, when considering open and dynamics MAS, it limits reusability and the agents' autonomy.

One of the main differences w.r.t. other works from the literature, e.g. [25,17,20], is that temporal constraints are defined in terms commitments and facts, which are, broadly speaking, the effects of the social actions, and not directly on actions (events). This improves flexibility and easiness of reuse in different contexts. Suppose, for instance, to have a specification with the action "pay-by-cash" with semantics *paid*, and to have the constraint "$paid\ before\ sent$". Then, suppose that a change in the context requires that also payment by credit cards can be performed. In this case it is necessary to add a new action "pay-by-credit-card" with the same semantics of "pay-by-cash", i.e. *paid*. Since the constraint is not defined directly on actions, it is not necessary to change it or add a new constraint. Indeed, $paid\ before\ sent$ already constrains the execution of both actions. These aspects are studied in [6], where the adaptation of the Contract Net Protocol to different contexts is discussed. However, where needed 2CL constraints can be used to rule directly actions. The way this is done is by adding a specific effect for each action and then by using these effects in the definition of the constraints.

Orders among actions could be obtained also by adding ad-hoc preconditions to the executability of actions, as done in [28,11,16,12]. However, this solution is not flexible, since regulations are hidden in the actions' definitions and thus difficult to be recognized, updated or modified. Moreover, agents should be always free to decide whether sticking to regulations. If regulations are realized by means of preconditions, agents cannot but choose which action to perform among those that are executable. Thus, they are forced to respect the rules and this is against the normative nature of regulations [13]. In our approach, which is orthogonal to preconditions definition, agents are free to evaluate different alternative paths and to take advantage from opportunities by choosing, among these, the most convenient for them. The role of the constraints is to restrict the set of legal executions, but an agent is free to decide to stick at the rules or to violate them. In the second case the agent knows it could be punished (sanctioned).

A real case study in which this approach has been tested is for the representation of *MiFID: Markets in Financial Instruments Directive* [8]. This directive by the European

Union regulates the interaction of banks, clients, and financial intermediaries in order to guarantee the investor from the intermediaries. The complete example can be found at `http://www.di.unito.it/~alice/2CL/`.

### 2.2 *Why (and How)* Committing to regulations

In many practical situations, commitments involve rich temporal structures rather than simple conditions to achieve. Let us consider a few examples: (a) an insurance company commits to reimbursing a covered patient for a health procedure provided the patient obtains approval from the company prior to the health procedure; (b) a pharmacy commits to provide medicine only if the patient obtains a prescription for that medicine; (c) an insurance company commits to paying an innetwork surgeon for a procedure only after a covered patient has undergone the procedure. Presumably, the surgeon would bill the insurance company after performing the procedure.

Commitments alone do not have the degree of expressiveness required by these conditions. Indeed, conditions in conditional commitments do not impose a temporal ordering: the consequent condition can be achieved even if the antecedent condition does not hold. The contribution of the thesis for capturing these aspects is described in [18] and consists in a new formalization of commitments, where temporal regulations are incorporated as content of commitments themselves. In this way regulations assume a normative force which is due to the regulative nature of commitments. For example, $C(x, y, \top, a\,before\,b)$ expresses the engagement of $x$ towards $y$ not only to make $a$ and $b$ happen but also to make them happen in the given order. Participants to the interaction will be able to guide their actions locally, in order to not violate any commitment they have taken, and to judge the compliance of their counter-parties. Indeed, since regulations are placed inside commitments, the debtor will be considered responsible and thus liable for violations. Consider, for example, a regulation saying that a physician's referral should precede a surgeon's procedure; in this situation, in case of violation, it is not clear whether the physician is responsible for moving first or the surgeon is responsible for moving second. By placing the regulations in commitments, we make it explicit that it is the debtor of the commitment who needs to ensure its satisfaction.

For this reason it becomes fundamental for an agent to establish, before taking a commitment, if it has a sufficient support by the other agents. The elements the agent has to consider are both the set of actions it can perform and the cooperation it can get from the others, via the set of commitments of which it is the creditor. To this aim, we formalized the notions of *control* and *safety*. The former captures the capability, for an agent, to bring about a regulation. It depends on the actions a certain agent can perform and on commitments directed towards it. The latter is strictly related to the notion of control: a commitment is safe if its debtor has established sufficient control to guarantee being able to discharge it.

To the best of our knowledge, no approach for protocol specification based on commitments allows to express commitments to temporal regulations. Placing temporal regulation inside commitments, however, allows for the representation of a debtor and thus allows to precisely identify who is responsible for each regulation and potentially liable for a violation. This is an advantage w.r.t. approaches based on expectations [1] which are not scoped by a debtor and a creditor. Moreover, it helps make the regulations explicit within the system of interacting agents and thereby facilitates their coordination. Accordingly to [13] it allows also to explicitly represent the recognition and the acceptance of a regulation by the agents.

## 3 Related Works

The need of expressing temporal regulations is supported by many attempts in the literature to rule actions' execution along the interaction. However, in our opinion, all these attempts can be improved in order to better take into account regulative aspects without compromising the flexibility and all the good properties of commitment protocols.

Fornara and Colombetti [14] propose a model based on *interaction diagrams*, a kind of specification which is similar to UML sequence diagrams. The choice of relying on interaction diagrams is very strong because it forces the ordering of action execution defining a strict set of allowed sequences. It basically can be classified as a procedural approach, thus presenting the same shortcomings [20]: it weakens agents' autonomy to decide which action to perform and their capability to take advantage of opportunities; it is too rigid, where instead the openness and dynamicity of MASs require higher flexibility of the specification.

The use of a declarative approach is proposed by Singh [25] and by Mallya and Singh [17]. In these works they define a *before relation* applied to events. The idea is that when a before relation among two activities is specified, the only thing that matters is the order among the two, no matter what happens in-between. Even if the choice of adopting a declarative specification overcomes many limits of the proposals described before, the main limitation is that temporal regulations are defined over actions (events). As described in the previous section, a greater degree of decoupling between actions and temporal specifications can, in our opinion, support better the openness of MAS. Moreover, this kind of regulation is conceived as a solution for service composition external from protocol specification. In our proposal, instead, temporal regulations actively contribute to the definition of the protocol.

The same shortcoming can be found in other proposals. It is hard to be exhaustive but let us consider a proposal inspired from the neighboring area of business processes. Pesic and van der Aalst [22] propose ConDec, a declarative language for business process representation. ConDec is a graphical language grounded in Linear-time Temporal Logic, which is used to rule the activities that compose a process. Montali and colleagues [9,20] integrate ConDec with SCIFF thus giving a semantics to actions that is based on expectations. The authors use this approach to specify interaction protocols and service choreographies. As the previous one, also this proposal is based on actions, thus suffering the same shortcomings.

Dialogue games are another solution for communication specification. Different kinds of dialogues basically define different kind of schema according which the agents can interact. The differences among them are given by the aim of the communication (e.g. persuade, inform, negotiate). Our approach is more general, since it provide the basic components for interaction specification and since it is not limited to communication (message exchange), as in [15,19], but to interaction in general. By means of this tool the desired interaction can be declaratively drawn according to the needs and to the aim of the system, without having to choose one among predefined schema. This is along the line of the claim by Singh in [10]. The idea is that a standard is difficult to be used "as-is" for modelling a desired system. To model a desired interaction he proposes standards for standard definition. Similar considerations holds for works that propose to use commitments for ACL semantics: they define predefined schema (type) for actions specification, bringing to an undesired rigidity. Our approach, and in general approaches based on the meaning of messages [10], are more flexible.

## 4 Ongoing and Future Work

Our proposal, for patterns specification and commitments to temporal regulations, allows facing two different lacks of commitment protocols related to temporal regulations. The thesis will finally investigate a unified framework in which both aspects can be reconciled under a common normative force. In other words, agents will be provided of the necessary means to explicitly recognize and accept temporal regulations, thus accepting their behaviour to be influenced by them [13]. Of course, agents will be free to decide to violate them in every moment. Thus, this framework will allow agents to commit to complex conditions, it will allow for the specification of patterns of interaction representing norms, conventions and rules, and it will provide the tools necessary to the agents to verify their ability to fulfill the engagements (along the line of control and of safety).

Reconciling these two aspects, which are *strictly connected to one another* [5], opens the way to interesting considerations. In particular, a set of constraints restricts the set of commitments that can be taken by the agents to those that can be considered legal. For example, before getting on a train a person has to punch the ticket. Only after, he/she is allowed to travel on the train. However, think to a person that commits to travel to his/her destination first and, once he/she reached it, to punch the ticket. In this situation, it is impossible for the person to fulfill his/her commitment without violating the norm. More generally, in order to propose a unified framework some important questions are to be answered. For example, given a set of norms expressed in terms of patterns of interaction, how can one establish which commitments are compliant and which are incompatible? If norms change, how do these changes affect the set of commitments? Moreover, how can one monitor the interaction of the agents and discover violations? In this respect, a solutions could be to lean on e-institutions. In [3] an initial proposal is described, where the idea is to define specific artifacts [21] able to detect violations. The kinds of reasoning that can be performed in this way are many. For example, it is possible not only to detect a violation, but also to classify different violations according to how relevant they are or how costly would be to repair from the damage caused by the violation.

The modularity of our proposal suits well the needs of the dynamic specification of protocols, along the line of [2]. Artikis' proposal is to define a set of *meta-actions* that can be performed by the agents at run-time, and that can change the set of rules that define the protocol. During this phase, the interaction is suspended. It will be resumed once the definition of the new rules is finished. In our proposal, it is possible to define a set of meta-actions whose effects are to change the set of constraints representing the norms that must be respected in the MAS. By performing those actions, agents would be able to change at run-time and dynamically, i.e. without suspending the interaction, the set of rules. Also this extension opens the way to some important question. For instance, who and how is allowed to change the rules? As part of the future work we will investigate also these aspects.

Finally, our proposal can be applied also to business process representations, and in particular to those situations in which a sequential representation is not adequate, due to the many alternative executions. In these contexts the high number of possible sequences suggests that a declarative representation, based on rules or constraints, is preferable with respect to procedural approaches. We plan to investigate more deeply these aspects along the line of [27], where a business process is described in terms of the commitments of the actors, that are involved in the process. One advantage of adopting declarative specifications and a modular representation of the constitutive (actions) and the regulative (constraints) part, is a gain of time and money in the operation of update of the business process due, for example, to norms changes.

## 5  Acknowledgments

We would like to thank Viviana Patti, Munindar P. Singh, and Amit K. Chopra for the helpful discussions and the work done together. We would like to thank also the reviewers for the helpful suggestions.